\documentclass[sigconf]{acmart}

\usepackage{booktabs} 
\usepackage{balance}  
\usepackage{graphicx} 
\usepackage{times}    
\usepackage{url}      
\usepackage{helvet}
\usepackage{courier}
\usepackage{amssymb}
\usepackage{amsmath}
\usepackage{amsbsy}
\usepackage{algorithm}
\usepackage{caption}
\usepackage[noend]{algpseudocode}
\usepackage{textcomp}
\usepackage{subfig}
\usepackage{color}

\setcopyright{none}

\begin{document}
	
	\title{Interactive Prior Elicitation of Feature Similarities for Small Sample Size Prediction}

	\author{Homayun Afrabandpey}
	\affiliation{%
		\institution{Helsinki Institute for Information Technology HIIT, Dept. of Computer Science, Aalto University}
	}
	\email{homayun.afrabandpey@aalto.fi}
	
	\author{Tomi Peltola}
	\affiliation{%
		\institution{Helsinki Institute for Information Technology HIIT, Dept. of Computer Science, Aalto University}
	}
	\email{tomi.peltola@aalto.fi}
	
	\author{Samuel Kaski}
	\affiliation{%
		\institution{Helsinki Institute for Information Technology HIIT, Dept. of Computer Science, Aalto University}
	}
	\email{samuel.kaski@aalto.fi}
	
	\begin{abstract}
		Regression under the ``small $n$, large $p$'' conditions, of small sample size $n$ and large number of features $p$ in the learning data set, is a recurring setting in which learning from data is difficult. With prior knowledge about relationships of the features, $p$ can effectively be reduced, but explicating such prior knowledge is difficult for experts. In this paper we introduce a new method for eliciting expert prior knowledge about the similarity of the roles of features in the prediction task. The key idea is to use an interactive multidimensional-scaling (MDS) type scatterplot display of the features to elicit the similarity relationships, and then use the elicited relationships in the prior distribution of prediction parameters. Specifically, for learning to predict a target variable with Bayesian linear regression, the feature relationships are used to construct a Gaussian prior with a full covariance matrix for the regression coefficients. Evaluation of our method in experiments with simulated and real users on text data confirm that prior elicitation of feature similarities improves prediction accuracy. Furthermore, elicitation with an interactive scatterplot display outperforms straightforward elicitation where the users choose feature pairs from a feature list.
	\end{abstract}
	
	%
	%
	
	
	\maketitle
	
	\section{Introduction}\label{SecI}
	
	Regression analysis becomes difficult when the sample size is substantially smaller than the number of features. ``Small $n$, large $p$'' refers to the generic class of such problems which arise in different fields of applied statistics such as personalized medicine \cite{costello2014community, tian2014simple} and text data analysis \cite{forman2003extensive, qu2010bag}. The problem poses several challenges to standard statistical methods \cite{johnstone2009statistical} and demands new concepts and models to cope with the challenges. An important challenge is that prediction by fitting regression models using traditional techniques is an ill-posed task in ``small $n$, large $p$'' and is unlikely to be accurate and reliable. Regularization methods \cite{tibshirani1996regression, zou2005regularization} have been proposed to cope with this challenge; however, the improvement they can give is limited. Additionally, modelling could use prior information, i.e. information available about the problem prior to observing the learning data. Prior information is often available only as the experience and knowledge of experts. The process of quantifying and extracting user's prior knowledge is known as prior elicitation. The extracted knowledge can be used to improve an underlying model. The two main questions in the process are how to quantify the prior knowledge, and how to plug-in the extracted prior knowledge to the model.
	
	Garthwaite et al \cite{garthwaite2013prior} proposed a method of defining the full prior distribution for a generalized linear model by quantifying experts' opinions on different statistics such as the median, lower and upper quantiles. Interactive Principal Component Analysis (iPCA) \cite{jeong2009ipca} supports data analysis of multivariate data sets through modification of the model parameters by the user. The drawback of these types of prior elicitation is that they assume users are experts in the underlying model and not just domain-experts. To solve this problem, observation-level interaction has been proposed where the focus is on interaction between the user and the data rather than model parameters \cite{brown2012dis, endert2011observation}. Using the extracted knowledge from the interaction, the parameters of the underlying model are tuned to reflect the user's knowledge. In recent work, Daee et al \cite{daee2016knowledge} proposed a method of eliciting user's knowledge on single features to improve the predictions in a sparse linear regression problem. The user's knowledge assumed to be about feature relevance and/or feature weight values. Similarly, Micallef et al \cite{micallef2016interactive} proposed an interactive visualization to extract user's knowledge on the relevance of individual features for a prediction task. 
	
	In this paper, we present a novel approach on interactive prior elicitation of pairwise similarities of features in ``small $n$, large $p$'' prediction task. The proposed approach uses an interactive MDS-type scatterplot of the features to let users give feedback on their pairwise similarities, in the sense of how similarly they would affect the predictions. Based on this input, the system learns a new similarity metric for the features and redraws the scatterplot. Finally, the learned metric is used to define a prior distribution for the prediction parameters. The proposed approach shields users from the technicalities of the underlying model.
	The contributions of this paper can be summarized as:
	\begin{itemize}
		\itemsep0em
		\item User's prior knowledge is quantified as the prior covariance of the regression coefficients in a Bayesian linear regression model. Using this interpretation, our system lets the user manipulate the prior distribution of the model parameter indirectly by his feedback, without having to understand modelling details.
		\item Feedback is collected on pairwise similarities of the features rather than the data, parameters or single features. This type of feedback is complementary to all earlier approaches.
		\item The prior is elicited with an MDS-type of interactive visualization that has earlier been used for visualizing similarities of data items.
	\end{itemize}
	Our simulation results and preliminary user study demonstrate that when collecting pairwise similarity knowledge using the proposed interactive intelligent interface, users are able to provide more informative feedback, and the performance of the underlying model increases in prediction tasks.
	
	\section{Overview} \label{SecII}
	
	To motivate our algorithm and for the purpose of clarity, we illustrate our basic idea with a simple use case. We used the sentiment data set \cite{blitzer2007biographies} which contains text reviews and the corresponding rating values (taken from \url{www.amazon.com}) of four product categories. Each review is represented using a vector of unigram and bigram keywords that appear in at least 100 reviews within the same category. We focus on the \textit{kitchen appliances} category where there are $5149$ reviews, each represented by a feature vector of size $824$ \cite{Hernandez2015Expectation}. The task is to learn a model that linearly relates the keywords (which here are features) to the ratings (outputs) to predict the ratings from the textual content of the reviews. This is a supervised learning task where we have a training set of inputs $\pmb{X} \in \mathbb{R}^{5149\times824}$ and outputs $\pmb{y} \in \mathbb{R}^{5149\times 1}$. To simulate the ``small $n$, large $p$'' paradigm, we randomly select 100 reviews and their corresponding ratings as the training set.
	
	A linear regression model for this task can be defined using a parameter vector $\pmb{\beta} = (\beta_1, \beta_2, ..., \beta_{824}) \in \mathbb{R}^{824\times 1}$. Mathematically, the model is
	\begin{equation}\label{EQ1}
		\pmb{y} = \pmb{X}\pmb{\beta} + \pmb{\epsilon}
	\end{equation}
	where $\pmb{\epsilon} \sim \mathcal{N}(0,\sigma_{noise}^2\pmb{I})$ is the residual noise. Equation \ref{EQ1} induces a Gaussian distribution for the likelihood as $\pmb{y}|\pmb{\beta},\sigma_{noise}^2 \sim \mathcal{N}(\pmb{X}\pmb{\beta},\sigma_{noise}^{2}\pmb{I})$. The goal is to learn the posterior distribution of $\pmb{\beta}$ given the training data.
	
	Inferring the posterior of the parameters in the Bayesian setting requires a prior distribution. In data sets with large sample sizes, the choice of the prior distribution will have a minor effect on the posterior inferences; however, since we assumed a ``small $n$, large $p$'' data set, the role of the prior distribution becomes more important. Setting prior distributions is a difficult task and requires knowledge on both the domain and the model parameters. In this paper, we introduce a method for helping in this task, by learning and refining a good prior distribution for the prediction parameters using feedback given by a user. User's knowledge is assumed to be about the pairwise similarities of the keywords with regard to the role they have in the prediction task. In other words, we mean that keywords have a similar effect on the rating values (the values of the regression coefficients are similar). As an example, keywords ``\textbf{good}'' and ``\textbf{excellent}'' have a similar role in the prediction since both of them convey information that the user will give a high rating to the product, while keywords ``\textbf{bad}'' and ``\textbf{good}'' are dissimilar.
	
\begin{figure}[]
		\centering
		\subfloat[]{
			\includegraphics[clip, width = 0.82\columnwidth]{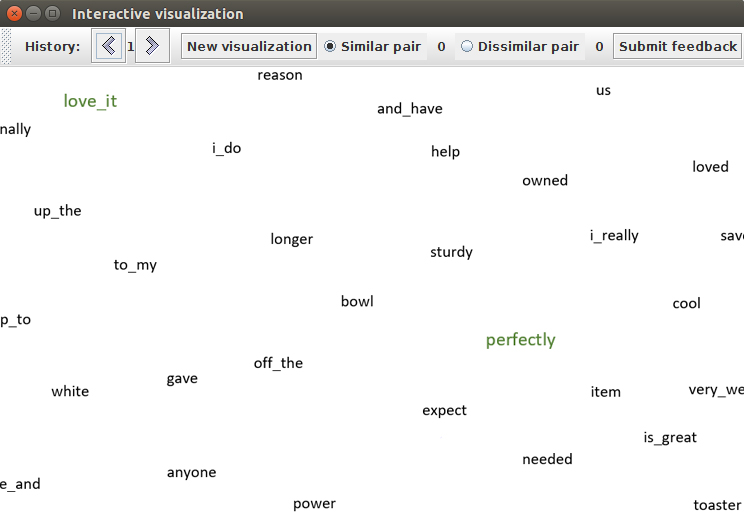}
			\label{fig1a}
		}
		
		\subfloat[]{
			\includegraphics[clip, width = 0.82\columnwidth]{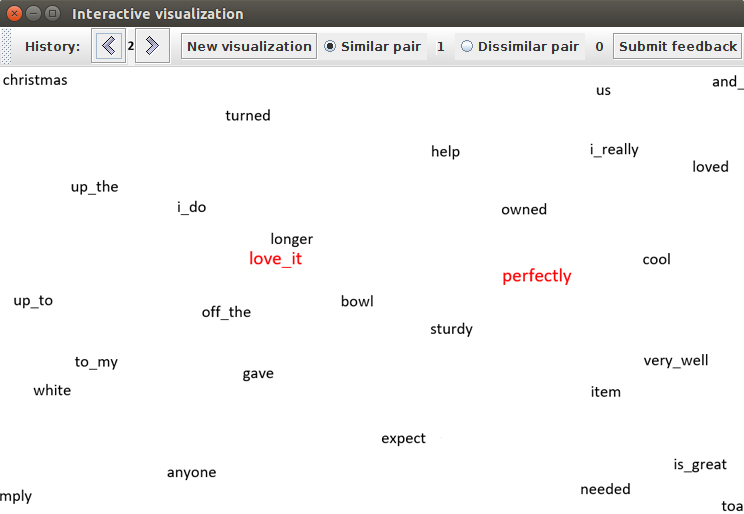}
			\label{fig1b}
		}
		\caption{The scatterplot (a) before submitting feedback, and (b) after submitting feedback and requesting a new visualization. In both figures, the scatterplot is zoomed in suitably to better show the keywords}
		\label{fig1}
	\end{figure}

	Figure \ref{fig1} illustrates an example interaction between the user and our system. Keywords (features) are visualized to the user on the scatter plot, where she can zoom in/out by scrolling down/up the mouse. The user investigates the distances among keywords and decides whether two keywords should be closer to each other (similar) or farther away from each other (dissimilar) based on her prior knowledge. As an example, the user concluded that according to her prior knowledge, the distances between keywords \textbf{``love\_it''} and \textbf{``perfectly''} should be less than what is shown in the scatterplot. She selects these keywords by clicking on them (their color will change to green as shown in Figure \ref{fig1a}), selecting similar/dissimilar box in the menu bar and then clicking on the submit button. Then the user can ask for a new visualization (\textbf{New Visualization} button in Figure \ref{fig1}) to see the effect of her feedback on the distances between keywords (Figure \ref{fig1b}), or she can continue giving more feedback according to current distances. As shown in Figure \ref{fig1b}, the one feedback given by the user modifies the distances between keywords, however it was not informative enough to make distances perfect. This will iterate until the user is satisfied with the visualization. The knowledge extracted from the user is used to build a proper covariance matrix for the prior distribution of the prediction parameter $\pmb{\beta}$. Finally, using the obtained prior, we compute the posterior of the prediction parameters.
	\setlength{\textfloatsep}{2pt}
	\section{Interactive Prior Elicitation of Pairwise Similarities} \label{SecIII}
	
	We reformulate the Interactive Neighbor Retrieval Visualizer \cite{peltonen2013information}, which visualized data items, to a method for prior elicitation on features. To visualize the features for the user, we use the original data space as the representation for the features in the high-dimensional space. More precisely, we define $\pmb{f}_{i} = [x_{1i}, \; \dots \; ,x_{ni}]^T$ as the original representation of the $i^{th}$ feature, where $x_{ni}$ is the $i^{th}$ element of the $n^{th}$ sample. With this definition, we have $D$ features, each of which is an $n$-dimensional vector. We define $\{\pmb{g}_i\}_{i=1}^D$ as the corresponding low-dimensional projections of $\{\pmb{f}_i\}_{i=1}^D$, to be learned from user feedback. 
	
	At each iteration $t$, we define the similarity matrix of the features in the high-dimensional space as
	\begin{equation}\label{EQ3}
		\pmb{P}^t = \Bigg [ p_{j|i}^t = \frac{exp(-\parallel \pmb{f}_{i} - \pmb{f}_{j} \parallel_{\pmb{A}^t}^2/\sigma_{i}^{2})}{\sum_{k \neq i}exp(-\parallel \pmb{f}_{i} - \pmb{f}_{j} \parallel_{\pmb{A}^t}^2/\sigma_{i}^{2})} \Bigg ]_{i,j = 1}^D
	\end{equation} 
	where $\pmb{A}^t$ is the unknown similarity metric between the features, $\parallel \pmb{f}_i - \pmb{f}_j\parallel_{\pmb{A}}^2=(\pmb{f}_i-\pmb{f}_j)^T \pmb{A} (\pmb{f}_i-\pmb{f}_j)$ and $\sigma_i^2$ is a scaling parameter. The unknown similarity metric $\pmb{A}^t$ encodes the user feedback and is learned iteratively by interaction with the user. The metric is initialized to unit matrix. 
	
	To find the location of the points in the visualization space at iteration $t$, an analogous matrix is defined for the low-dimensional projections:
	\begin{equation}\label{EQ4}
		\pmb{Q}^t = \Bigg [ q_{j|i}=\frac{exp(-\parallel \pmb{g}_{i}^{t} - \pmb{g}_{j}^{t}\parallel^2/\sigma_i^2)}{\sum_{k\neq i}exp(-\parallel \pmb{g}_{i}^{t} - \pmb{g}_{k}^{t}\parallel^2/\sigma_i^2)} \Bigg ]_{i,j = 1}^D.
	\end{equation}
	Finally, the locations of the points in the low-dimensional space are obtained by optimizing the following expected cost function \cite{peltonen2013information}:
	\begin{equation}\label{EQ5}
		\mathbb{E}[C] = \mathbb{E}_{\pmb{A}|F}[\lambda \mathbb{E}_i[KL(P_i,Q_i)]+(1-\lambda)\mathbb{E}_i[KL(Q_i,P_i)]],
	\end{equation}
	where $\mathbb{E}_{\pmb{A}|F}$ denotes the expectation over the posterior distribution of the learned metric given the feedbacks $F$, and $\mathbb{E}_i$ is expectation over the training set points. Since the high-dimensional distributions $P_i$ are functions of the unknown metric $\pmb{A}$, the cost function is represented as the expectation over the possible metrics. 
	The parameter $\lambda \in [0,1]$ controls the relative importance of recall and precision of the display \cite{venna2010information}. 
	The final similarity metric $\pmb{A}^{final}$, learned in the last iteration of user interaction, is used to define a prior distribution for the regression weights according to equations \ref{EQ6} and \ref{EQ7}:
	\begin{equation}\label{EQ6}
		\pmb{C} = \Bigg [ c_{ij} = \exp(-\frac{\parallel \pmb{f}_{i} - \pmb{f}_{j} \parallel_{\pmb{A}^{final}}^2}{2\sigma^2}) \Bigg ]_{i,j=1}^D,
	\end{equation}
	\begin{equation}\label{EQ7}
		\pmb{\beta} \sim \mathcal{N}(\pmb{0}, \sigma_{noise}^2 \tau^2 \pmb{C}),
	\end{equation}
	where $\sigma$ and $\tau$ are scalar scale parameters. In our implementation, the value of $\sigma$ is set by cross-validation.
	
	By defining this prior distribution for the regression coefficients, and gamma prior distributions on $\tau^{-2}$ and $\sigma_{noise}^{-2}$, the posterior distribution is analytically intractable, but can be efficiently approximated using Variational Bayes (e.g., \cite[Chapter~10]{Bishop2006}). This gives a Gaussian posterior approximation for $\pmb{\beta}$. Finally, the prediction is done using the posterior mean. Pseudocode of the proposed method is presented in Algorithm 1.
	\begin{algorithm}
		\caption*{Interactive Prior Elicitation Pseudocode}
		\begin{algorithmic}[1]
			\State Set $\pmb{A}^0 = \pmb{I}$ and $t = 0$.
			\While{ user gives more feedback }
			\begin{itemize}
				\item[b.] Optimize the cost function \ref{EQ5} using the metric $\pmb{A}^t$ and find the position of the features in the low-dimensional space, $[\pmb{g}_i^t]_{i=1}^D$, at iteration $t$.
				\item[c.] Ask the user to give feedback about the similarity of the role of the features.
				\item[d.] Set $t = t+1$.
				\item[e.] Learn the new metric $\pmb{A}^t$ using the method introduced in \cite{yang2007bayesian} and the user feedback.
			\end{itemize}
			\EndWhile
			\State Compute the matrix $\pmb{C}$ using $\pmb{A}^{final}$ (Eq. \ref{EQ6}) and define a prior distribution for the weights as $\beta \sim \mathcal{N}(\pmb{0}, \sigma_{noise}^2 \tau^2 \pmb{C})$.
			\State Compute the posterior of the weights and use that to predict output for a new sample.
		\end{algorithmic}
	\end{algorithm}
	\section{Simulation Experiment} \label{SecIV}
	
	We conducted a simulated study on the data set introduced in Section 2 with two scenarios where a simulated user (i) gives all feedbacks at once, and (ii) gives feedback sequentially. As baselines, we used Bayesian linear regression with unit prior covariance and Bayesian linear regression with the prior covariance used in the first round of our method ("Without Feedback" in the following, since the prior is obtained by setting $\pmb{A} = \pmb{I}$ and without using feedback). We used a set of 3149 randomly selected reviews with their corresponding ratings to construct the simulated user. This is done by using the mean of the posterior distribution of the regression coefficient vector of a Bayesian linear regression model trained on the randomly selected data. The simulated user assumes two similarity clusters: (i) features with the highest 30 regression coefficients and (ii) features with the lowest 30 regression coefficients. Features in these two clusters are dissimilar to each other. Since there are enough samples (3149) compared to the dimensionality of the data (824), the posterior mean of the regression coefficient is a good representative of the true values of the feature weights and consequently the similarity of the role of the features in the prediction task.
	
\begin{figure}[h!]
		\subfloat[]{
			\includegraphics[clip, width = 0.45\columnwidth]{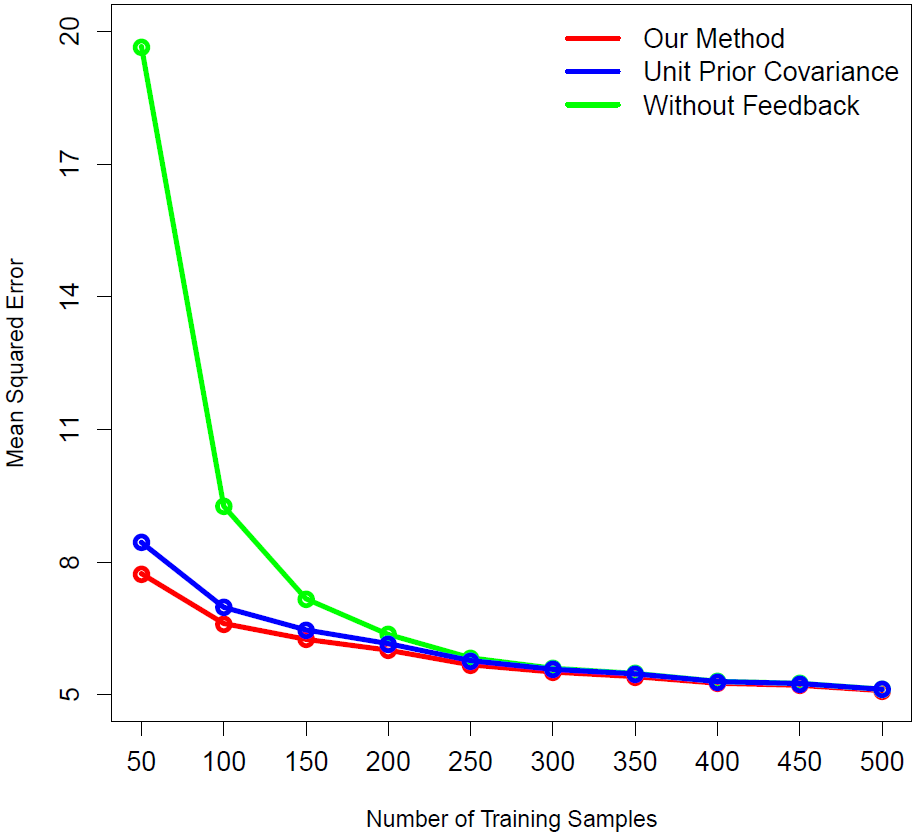}
			\label{fig2a}
		}	
		\subfloat[]{
			\includegraphics[clip, width = 0.47\columnwidth]{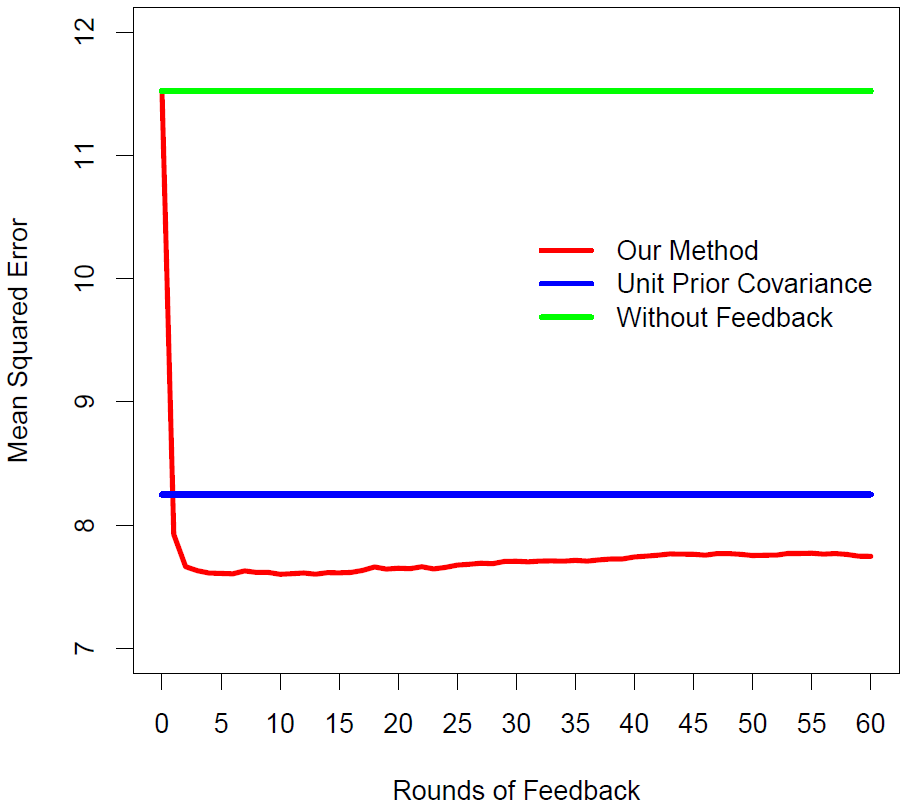}
			\label{fig2b}
		}
		\caption{Simulation results for (a) batch feedback, (b) sequential feedback.}
		\label{fig2}
	\end{figure}

	The remaining samples 
	are randomly partitioned into training and test sets. The results reported in this section are averaged over 10 simulated user construction iterations and 50 random training data selection. Figure \ref{fig2a} shows simulation results for the first scenario, in which the proposed method is evaluated with an increasing number of randomly selected training samples, from 50 to 500. Figure \ref{fig2b} shows the changes of Mean Squared Errors (MSE) on the test data with 100 randomly selected training samples when the simulated user gives feedback sequentially in 60 rounds; round 0 works without feedback
	. The simulated user gives 10 similarity feedback and 10 dissimilarity feedback in each round.
	
	From Figure \ref{fig2}, it can be concluded that assuming pairwise similarity/dissimilarity knowledge from the user, the proposed method improves the predictions by extracting prior knowledge.
	\setlength{\intextsep}{2pt}
	\section{User Study}
	
	We conducted a user study on 10 naive university students to empirically evaluate our two hypotheses that (i) by collecting prior knowledge on the pairwise similarity of the features we can improve predictions, and (ii) the interactive interface helps users to give better feedback and consequently improves the system's predictions. To evaluate the first hypothesis, we consider the same baselines used in the previous section. To evaluate the second hypothesis, we implemented two different versions of our system, both with the same underlying model, but with different interfaces: the proposed interactive interface and a simple non-interactive list visualization of the features. In the list visualization, the order of the features is random and fixed during the whole experiment for a user. The user goes through the list and selects the pairs which are similar or dissimilar according to her prior knowledge and gives feedback on them. This very simple interface was designed for testing hypothesis (ii). As far as we know, there are no earlier methods for the same task.
	
	We designed a between-subject study, where each participant performed two prior elicitation tasks with different interfaces and different data collections: the sentiment data set introduced in Section 2 and the reviews from the Yelp data set challenge (\url{www.yelp.com/dataset_challenge}). Users were asked to give feedback on pairwise similarity/dissimilarity of the words in the role they have in the prediction.
	For the Yelp data, we used a subset with 4086 reviews. In both data sets, we set a threshold on the tf-idf values (a standard technique in information retrieval, see \cite{sparck1972statistical}) of the words to choose 300 words. To simulate a ``small $n$, large $p$'' training data, we randomly selected five subsets of each data set with 100 samples, and used the rests for test. Therefore, the training set for each task contains 100 samples and 300 features. Each of the selected five subsets (from each data set) was used once for the interactive interface and once for the non-interactive interface with different users. Users interact with each interface for 20 rounds and give 5 feedbacks (similarity/dissimilarity) per round. The study was balanced with respect to the combination of the type of interface, task and order. After both tasks, a short semi-structured interview was conducted with each participant.
	
\begin{figure}[h!]
		\subfloat[]{
			\centering
			\includegraphics[clip, width = 0.45\columnwidth]{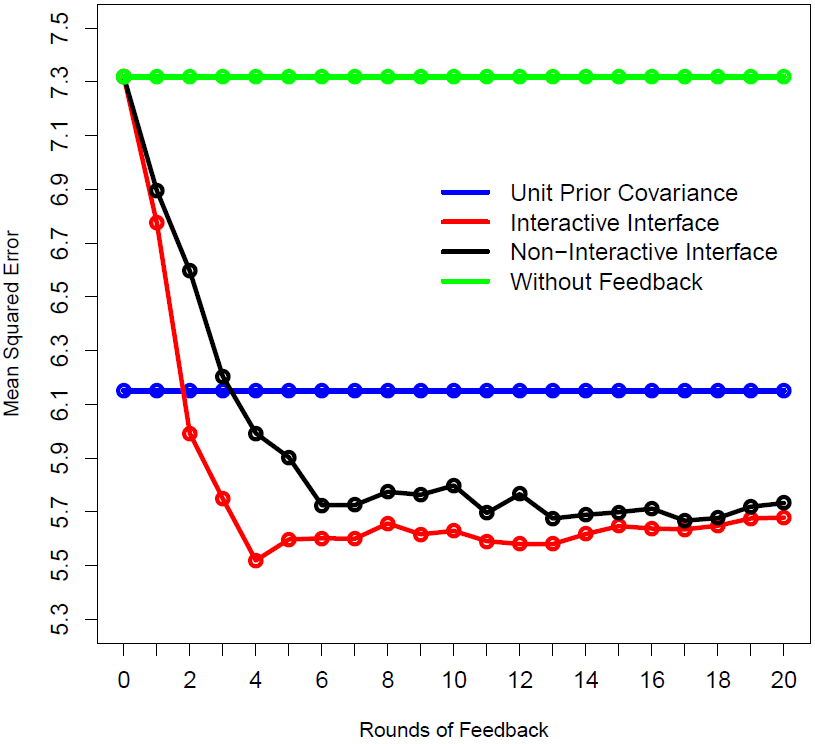}
			\label{fig3a}
		}	
		\subfloat[]{
			\centering
			\includegraphics[clip, width = 0.45\columnwidth]{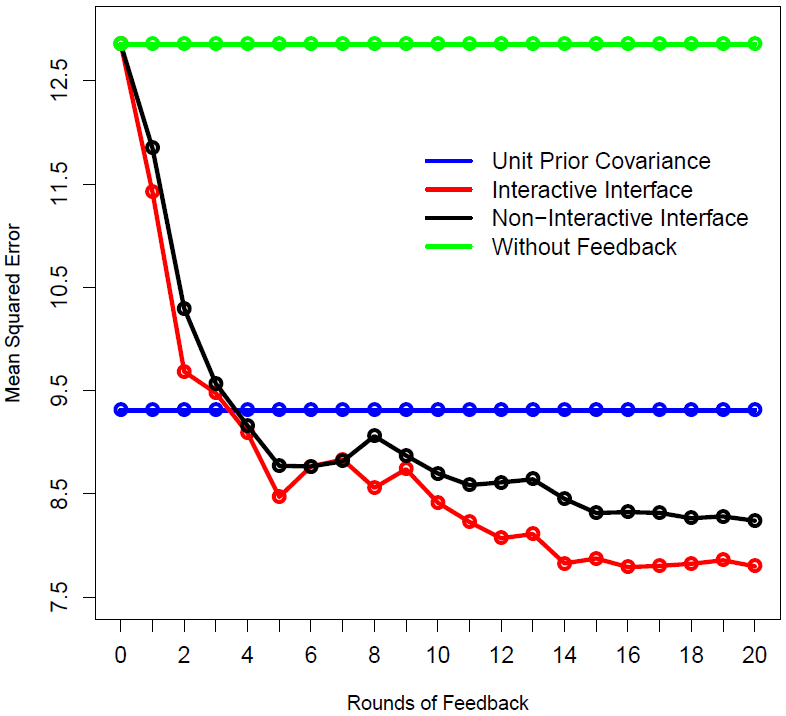}
			\label{fig3b}
		}
		\caption{Changes of prediction MSE by increasing number of feedbacks for (a) the subset of the sentiment data set and (b) the subset of the Yelp data set.}
		\label{fig3}
	\end{figure}

	Figures \ref{fig3a} and \ref{fig3b} show MSEs on test data as a function of the number of feedback iterations for the two data sets. 
	According to the figures, extracted prior knowledge of the user improves the mean squared errors of the predictions compared to both baselines. Moreover, the difference between the MSE values obtained by the interactive interface and the non-interactive interface shows the amount of improvement made to the predictions using the interface. To test the statistical significance of the improvements made by our method compared to each of the other methods, we used the same procedure introduced in \cite{micallef2016interactive}. The distance between the average curves in the last round (round $20$ in Figure \ref{fig3}) is used as the test statistics. By assuming that there is no difference between the results obtained by the interactive interface and other methods, we compute the distribution of the test statistics by performing $10^5$ permutations of the labels, e.g. interactive interface, non-interactive interface, etc. Finally, the proportion of the permutations which has higher values of the test statistics compared to the test statistics when using true labels, is used as $p$-value of the significance test. Based on this test, the improvement made on the ''Unit Prior Covariance'' ($p = 0.048$ for the sentiment and $p = 0.016$ for the YELP data set) and the ''Without Feedback'' ($p = 0.0$ for the sentiment and $p = 0.0$ for the YELP data set) baselines in both data sets are statistically significant, while for the non-interactive interface, the differences between MSEs are not statistically significant ($p = 0.73$ for the sentiment and $p = 0.56$ for the YELP data set) which might be due to the small number of users.
	
	It should be noted that since only a small portion of the words are meaningfully related to the rate prediction task, i.e. most of the words are verbs (am, is, etc.) or subjects (I, he, etc.) which are difficult for the user to give feedback on, users gave their best feedback in the first couple of rounds which causes prior elicitation to best improve the prediction errors in the first half of the rounds. But, in the second half of the rounds, prediction errors either improve slowly because of the repetitive feedbacks given by the user, or even sometimes drop since some users started to give feedback on irrelevant words.
	\setlength{\intextsep}{2pt}
	
	In the interview, 8 out of 10 users reported that they felt the intelligent interface helped them to accomplish the task. However, 5 users stated that they preferred the simple interface over the intelligent one. This is not surprising since people often prefer simpler systems over more complex ones \cite{hearst2009search} even if the complex system benefits them in accomplishing the required task.
	\section{Discussion and Conclusion}
	In this paper, we presented a new method and a prototype implementation of an interactive prior elicitation system which elicits an expert's prior knowledge on feature similarities to improve prediction accuracy. The system involves an intelligent user interface which helps the user in the interaction. We believe that this is an important step toward more efficient interactive prior elicitation methods. The main novelties are the type of feedback assumed from the user and the interpretation of the extracted knowledge as prior covariance for the parameter of the linear regression model.
	
	In the current implementation, we pruned the number of features to avoid overwhelming the user; however, for the general case of a large number of features, we are working on an active learning version of the method to prioritize the feature pairs and allow scaling up to a much larger number of features.

	\bibliographystyle{acm-sigchi}
	\bibliography{References}

\begin{thebibliography}{10}

\bibitem{Bishop2006}
Bishop, C.~M.
\newblock {\em Pattern Recognition and Machine Learning}.
\newblock Springer, 2006.

\bibitem{blitzer2007biographies}
Blitzer, J., Dredze, M., and Pereira, F.
\newblock Biographies, bollywood, boom-boxes and blenders: Domain adaptation
  for sentiment classification.
\newblock In {\em ACL}, vol.~7 (2007), 440--447.

\bibitem{brown2012dis}
Brown, E.~T., Liu, J., Brodley, C.~E., and Chang, R.
\newblock Dis-function: Learning distance functions interactively.
\newblock In {\em Proceedings of the IEEE Conference on Visual Analytics
  Science and Technology (VAST)} (2012), 83--92.

\bibitem{costello2014community}
Costello, J.~C., Heiser, L.~M., Georgii, E., G{\"o}nen, M., Menden, M.~P.,
  Wang, N.~J., Bansal, M., Hintsanen, P., Khan, S.~A., Mpindi, J.-P., et~al.
\newblock A community effort to assess and improve drug sensitivity prediction
  algorithms.
\newblock {\em Nature Biotechnology 32}, 12 (2014), 1202--1212.

\bibitem{daee2016knowledge}
Daee, P., Peltola, T., Soare, M., and Kaski, S.
\newblock Knowledge elicitation via sequential probabilistic inference for
  high-dimensional prediction.
\newblock {\em arXiv preprint arXiv:1612.03328\/} (2016).

\bibitem{endert2011observation}
Endert, A., Han, C., Maiti, D., House, L., and North, C.
\newblock Observation-level interaction with statistical models for visual
  analytics.
\newblock In {\em Proceedings of the IEEE Conference on Visual Analytics
  Science and Technology (VAST)} (2011), 121--130.

\bibitem{forman2003extensive}
Forman, G.
\newblock An extensive empirical study of feature selection metrics for text
  classification.
\newblock {\em Journal of Machine Learning Research 3\/} (2003), 1289--1305.

\bibitem{garthwaite2013prior}
Garthwaite, P.~H., Al-Awadhi, S.~A., Elfadaly, F.~G., and Jenkinson, D.~J.
\newblock Prior distribution elicitation for generalized linear and
  piecewise-linear models.
\newblock {\em Journal of Applied Statistics 40}, 1 (2013), 59--75.

\bibitem{hearst2009search}
Hearst, M.
\newblock {\em Search user interfaces}.
\newblock Cambridge University Press, 2009.

\bibitem{Hernandez2015Expectation}
Hern{\'a}ndez-Lobato, J.~M., Hern{\'a}ndez-Lobato, D., and Su{\'a}rez, A.
\newblock Expectation propagation in linear regression models with
  spike-and-slab priors.
\newblock {\em Machine Learning 99}, 3 (2015), 437--487.

\bibitem{jeong2009ipca}
Jeong, D.~H., Ziemkiewicz, C., Fisher, B., Ribarsky, W., and Chang, R.
\newblock {iPCA}: An interactive system for {PCA}-based visual analytics.
\newblock In {\em Computer Graphics Forum}, vol.~28, Wiley Online Library
  (2009), 767--774.

\bibitem{johnstone2009statistical}
Johnstone, I.~M., and Titterington, D.~M.
\newblock Statistical challenges of high-dimensional data.
\newblock {\em Philosophical Transactions of the Royal Society of London A:
  Mathematical, Physical and Engineering Sciences 367}, 1906 (2009),
  4237--4253.

\bibitem{micallef2016interactive}
Micallef, L., Sundin, I., Marttinen, P., Ammad-ud din, M., Peltola, T., Soare,
  M., Jacucci, G., and Kaski, S.
\newblock Interactive elicitation of knowledge on feature relevance improves
  predictions in small data sets.
\newblock {\em arXiv preprint arXiv:1612.02487\/} (2016).

\bibitem{peltonen2013information}
Peltonen, J., Sandholm, M., and Kaski, S.
\newblock Information retrieval perspective to interactive data visualization.
\newblock {\em EuroVis-Short Papers\/} (2013), 49--53.

\bibitem{qu2010bag}
Qu, L., Ifrim, G., and Weikum, G.
\newblock The bag-of-opinions method for review rating prediction from sparse
  text patterns.
\newblock In {\em Proceedings of the 23rd International Conference on
  Computational Linguistics}, Association for Computational Linguistics (2010),
  913--921.

\bibitem{sparck1972statistical}
Sparck~Jones, K.
\newblock A statistical interpretation of term specificity and its application
  in retrieval.
\newblock {\em Journal of documentation 28}, 1 (1972), 11--21.

\bibitem{tian2014simple}
Tian, L., Alizadeh, A.~A., Gentles, A.~J., and Tibshirani, R.
\newblock A simple method for estimating interactions between a treatment and a
  large number of covariates.
\newblock {\em Journal of the American Statistical Association 109}, 508
  (2014), 1517--1532.

\bibitem{tibshirani1996regression}
Tibshirani, R.
\newblock Regression shrinkage and selection via the lasso.
\newblock {\em Journal of the Royal Statistical Society. Series B
  (Methodological)\/} (1996), 267--288.

\bibitem{venna2010information}
Venna, J., Peltonen, J., Nybo, K., Aidos, H., and Kaski, S.
\newblock Information retrieval perspective to nonlinear dimensionality
  reduction for data visualization.
\newblock {\em Journal of Machine Learning Research 11\/} (2010), 451--490.

\bibitem{yang2007bayesian}
Yang, L., Jin, R., and Sukthankar, R.
\newblock Bayesian active distance metric learning.
\newblock In {\em Proceedings of the Twenty-Third Conference on Uncertainty in
  Artificial Intelligence}, AUAI Press (2007), 442--449.

\bibitem{zou2005regularization}
Zou, H., and Hastie, T.
\newblock Regularization and variable selection via the elastic net.
\newblock {\em Journal of the Royal Statistical Society: Series B (Statistical
  Methodology) 67}, 2 (2005), 301--320.

\end{thebibliography}
\end{document}